\documentclass{article}

\usepackage[utf8]{inputenc} 
\usepackage[T1]{fontenc}    
\usepackage{hyperref}       
\usepackage{url}            
\usepackage{booktabs}       
\usepackage{amsfonts}       
\usepackage{nicefrac}       
\usepackage{microtype}      
\usepackage{xcolor}         

\usepackage{wrapfig}

\usepackage{tcolorbox}
\usepackage{adjustbox}
\usepackage[numbers]{natbib} 
\usepackage{xspace}
\usepackage{multirow}
\usepackage{lipsum}
\usepackage{amsmath}
\usepackage{graphicx}
\usepackage{tabularx}  

\usepackage[preprint]{neurips_2025}
\usepackage{tcolorbox}

\usepackage{enumitem}
\setlist[itemize]{leftmargin=*, itemindent=0pt, itemsep=2pt, topsep=2pt}
\setlist[enumerate]{leftmargin=*, itemsep=2pt, topsep=2pt}
\setlist[description]{leftmargin=*, itemsep=2pt, topsep=2pt}

\newcommand{\method}{{GLIP-OOD}\xspace}

\title{\method: Zero-Shot Graph OOD Detection with Graph Foundation Model}

\author{
\textbf{Haoyan Xu$^{1,*}$, 
Zhengtao Yao$^{1,*}$,
Xuzhi Zhang$^1$,
Ziyi Wang$^2$,}\\
\textbf{Langzhou He$^3$,
Yushun Dong$^4$, 
Philip S. Yu$^3$,
Mengyuan Li$^1$, 
Yue Zhao$^1$} \\
$^1$University of Southern California \quad
$^2$University of Maryland, College Park \quad \\
$^3$University of Illinois Chicago \quad
$^4$Florida State University \\
\texttt{haoyanxu@usc.edu, zyao9248@usc.edu, xuzhizha@usc.edu,} \\
\texttt{ zoewang@umd.edu, lhe24@uic.edu,  yushun.dong@fsu.edu,} \\
\texttt{psyu@uic.edu, mengyuanli@usc.edu, yzhao010@usc.edu}\\
}

\begin{document}

\maketitle
\begingroup
\renewcommand\thefootnote{*}
\footnotetext{Equal contribution.}
\endgroup

\begin{abstract}
Out-of-distribution (OOD) detection is critical for ensuring the safety and reliability of machine learning systems, particularly in dynamic and open-world environments. 
In the vision and text domains, zero-shot OOD detection—which requires no training on in-distribution (ID) data—has advanced significantly through the use of large-scale pretrained models, such as vision-language models (VLMs) and large language models (LLMs). However, zero-shot OOD detection in graph-structured data remains largely unexplored, primarily due to the challenges posed by complex relational structures and the absence of powerful, large-scale pretrained models for graphs.
In this work, we take the \textit{first} step toward enabling zero-shot graph OOD detection by leveraging a graph foundation model (GFM). Our experiments show that, when provided only with class label names for both ID and OOD categories, the GFM can effectively perform OOD detection—often surpassing existing ``supervised'' OOD detection methods that rely on extensive labeled node data.
We further address the practical scenario in which OOD label names are not available in real-world settings by introducing \method, a framework that uses LLMs to generate semantically informative pseudo-OOD labels from unlabeled data. These generated OOD labels allow the GFM to better separate ID and OOD classes, facilitating more precise OOD detection—all \emph{without} any labeled nodes (only ID label names). To our knowledge, this is the first approach to achieve node-level graph OOD detection in a fully zero-shot setting, and it attains performance comparable to state-of-the-art supervised methods on four benchmark text-attributed graph datasets.
\end{abstract}

\section{Introduction}
Out-of-distribution (OOD) detection \cite{yang2024generalized, liu2020energy, hendrycks2016baseline} is the task of identifying inputs that deviate significantly from the training distribution. As machine learning models are increasingly deployed in dynamic and safety-critical environments, the ability to reliably detect OOD instances has become important for maintaining robustness, improving trustworthiness, and preventing catastrophic failures. 
Applications span many domains, including medical diagnosis \cite{DBLP:conf/nips/UlmerMC20}, autonomous driving \cite{gardille2023towards}, video analytics \cite{DBLP:journals/corr/abs-2503-06166}, and finance \cite{caron2022shortcut}, where unseen or unexpected data is common and consequential.

\noindent
\textbf{Graph OOD Detection and Current Landscape.}
Recently, OOD detection has gained traction in graph-structured data (known as graph OOD detection \cite{wu2023energy, marevisiting, xu2025few, song2022learning}), where the goal is to detect anomalous or previously unseen nodes, edges, or subgraphs. 
It is important to the domains such as social networks \cite{xiao2020timme}, citation graphs, and e-commerce platforms, where new entities (e.g., emerging users, novel research topics, or innovative products) may appear during inference. 
Node-level graph OOD detection methods generally require a substantial number of labeled in-distribution (ID) nodes to train a graph neural network (GNN)-based classifier, which subsequently produces class probability distributions (logits) for all nodes in the graph \cite{wu2023energy, marevisiting, xu2025few}. 
OOD detection is then performed in a post-hoc manner by applying scoring functions \cite{qin2025metaood}—such as energy-based methods \cite{wu2023energy, liu2020energy}, maximum softmax probability \cite{hendrycks2016baseline}, or uncertainty-based metrics \cite{marevisiting}—to these logits. Nodes with high energy or entropy scores are considered likely to be OOD samples.
However, these methods are fundamentally limited by their reliance on labeled ID data, which can be expensive and impractical to obtain in real-world graph scenarios. This raises a natural question: C\textit{an we perform graph OOD detection without relying on labeled nodes at all}?

\begin{wrapfigure}{R}
{0.46\textwidth} 
    \centering
    \vspace{-0.9cm}
    \includegraphics[width=1\linewidth]{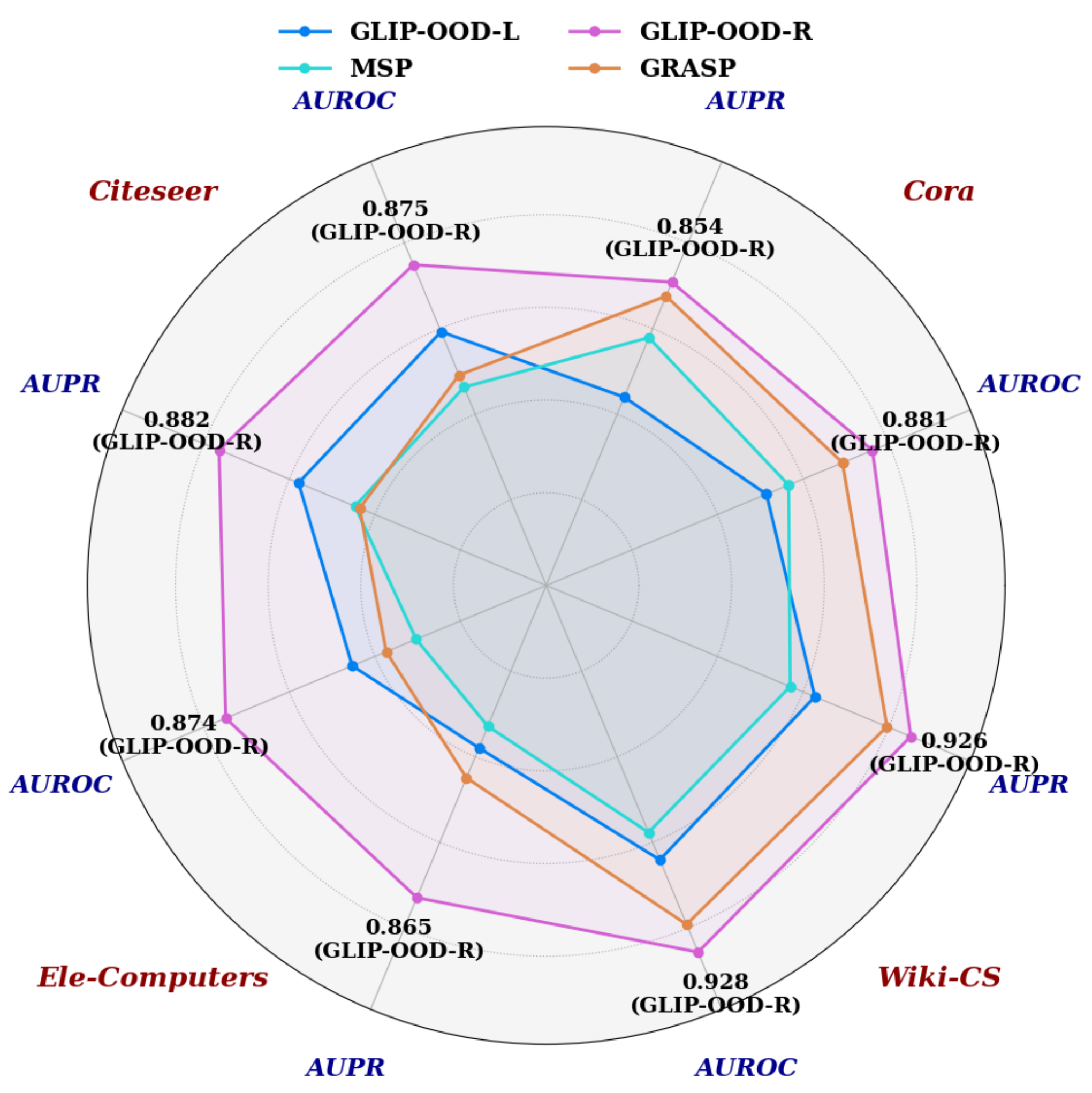}
    \vspace{-0.3in}
    \caption{Radar plot (AUROC and AUPR) on four text-attributed graph datasets. 
    \method-R uses real OOD class names, while \method-L employs LLM-generated pseudo-OOD names. 
    With no node-level supervision, \method-R exceeds supervised OOD baselines, and \method-L remains competitive even without real OOD labels.}
    \label{fig:Intro_img}
    \vspace{-0.5cm}
\end{wrapfigure}

\noindent
\textbf{Zero-Shot OOD Detection in Vision and Text.}
In computer vision and NLP areas, zero-shot OOD detection—which requires no training on ID data—has emerged as a compelling research direction. In the image domain, methods \cite{wang2023clipn, ming2022delving, esmaeilpour2022zero, cao2024envisioning,liu2024cood} leverage the capabilities of large-scale
vision-language models (VLMs), e.g., CLIP \cite{radford2021learning}, to detect OOD samples across a variety of ID datasets without requiring any labeled training examples. For example, CLIPN \cite{wang2023clipn} enhances CLIP with the ability to distinguish between OOD and ID samples by leveraging both positive semantic prompts and negation-based prompts, enabling zero-shot OOD detection in the image domain. ZOC \cite{esmaeilpour2022zero} extends CLIP by dynamically generating candidate unseen labels per test image and introducing a confidence score based on feature-space similarity to both seen and generated unseen labels.
In the text domain, large language models (LLMs) have demonstrated impressive zero-shot capabilities by leveraging natural language prompts and contextual understanding to detect OOD content \cite{zhang2024your, jin2022towards}. 
The ability to perform effectively with limited data makes LLM-based OOD detection particularly attractive for real-world applications.
Despite these advances, zero-shot OOD detection in graph domains remains largely unexplored due to the unique challenges posed by relational structures. More importantly, until now, there has been no widely acknowledged, powerful, large-scale foundation model—like VLMs and LLMs—for graphs.

\noindent
\textbf{Our Proposal: Zero-Shot OOD Detection in Graphs (\method).}
In this paper, we take the first step toward zero-shot graph OOD detection using a graph foundation model (GFM) \cite{zhu2024graphclip, liu2023towards, liu2025graph, chen2024text}. 
Similar to CLIP, the GFM does not inherently support OOD detection and instead matches any given node to one of the provided class labels. 
Below, we describe two scenarios:
\textit{\textbf{1. \method-R (Ideal Scenario: All Class Names).}
}When all class names (ID and OOD) are known, we supply only these label names to the GFM and derive the output logits to perform zero-shot OOD detection. 
Figure~\ref{fig:Intro_img} shows that, surprisingly, with only these label names (\method-R), the GFM already surpasses many supervised methods that rely on numerous ID-labeled nodes.  
This result highlights the GFM's strong intrinsic capacity for zero-shot OOD detection, given clearly defined ID and OOD classes.

\textit{\textbf{2. \method-L (Practical Scenario: ID Names + Pseudo-OOD).}}
More practically, the OOD classes are often unknown during training.
We thus generate \emph{pseudo}-OOD labels from unlabeled nodes to construct a pseudo-OOD label set alongside the known ID labels.
Concretely, we sample a small subset of unlabeled nodes, ask an LLM to check if each node’s text matches any known ID category, and, if not, treat it as pseudo-OOD. 
The LLM then summarizes the node to produce a plausible OOD label name. 
This procedure yields new pseudo-OOD categories, which we feed back into the GFM for zero-shot OOD detection. 
Our approach requires no real OOD labels or node annotations, yet it can assign nodes to either known ID or newly synthesized OOD classes—enabling fine-grained OOD detection. 
As shown in Figure~\ref{fig:Intro_img} (\method-L), the resulting performance is often comparable to fully supervised methods despite using no node-level supervision.

\noindent
We summarize our key contributions as follows:
\begin{itemize}[leftmargin=*, itemindent=0pt, itemsep=2pt]
\item \textbf{Novel Setting.} We are the first to study the problem of \emph{zero-shot} graph OOD detection, providing a comprehensive setting and framework that detects OOD nodes without any node-level labels.

\item \textbf{Zero-shot Solution.} We propose \method, which leverages LLMs to generate pseudo-OOD labels, semantically distant from known ID labels but close to real OOD distributions in the transductive graph setting, then uses a GFM in an augmented label space for OOD detection.

\item \textbf{Effectiveness.} Across four benchmark text-attributed graph (TAG) datasets, our zero-shot method outperforms supervised approaches (when given all label names) and achieves performance comparable to those methods using only ID labels plus pseudo-OOD labels. 
\end{itemize}

\section{Preliminary}
Our study focuses on OOD detection on TAGs. A TAG is represented as $G_T = (\mathcal{V}, \mathbf{A}, \mathcal{M}, \mathbf{X})$, where the set of nodes is $\mathcal{V} = \{v_1, \dots, v_n\}$, and each node is associated with raw text. These text attributes $\mathcal{M} = \{m_1, m_2, \dots, m_n\}$ are transformed into embeddings $\mathbf{X} = \{x_1, x_2, \dots, x_n\}$ using a pretrained model~\cite{reimers2019sentence}. The adjacency matrix $\mathbf{A} \in \{0, 1\}^{n \times n}$ encodes the graph's connectivity, where $\mathbf{A}[i, j] = 1$ indicates an edge between nodes $v_i$ and $v_j$.

\noindent \textbf{Node-level Graph OOD Detection}.
We have a labeled node set $\mathcal{V}_I$ and a large unlabeled node set $\mathcal{V}_U$. Each labeled node belongs to one of $K_{\text{ID}}$ known classes $ \mathcal{Y}_{\text{ID}} = \{y_k\}_{k=1}^{K_{\text{ID}}}$, while an unlabeled node may belong to an unknown class not included in $\mathcal{Y}_{\text{ID}}$. 
The problem is framed in a semi-supervised, transductive setting, where we can access the full set of nodes during training but only a subset of class labels (ID classes).
Our goal is to determine, for each node $v_i \in \mathcal{V}_U$, whether it is from the $K_{\text{ID}}$ ID classes or from the OOD unknown classes.

\noindent \textbf{Zero-shot Graph OOD Detection with All Labels}.
We provide the model only with the ID and OOD class names and do not assume access to any node labels in the graph. That is, all nodes in the graph are unlabeled, and the model directly outputs an OOD score for each input node based solely on the label names, node attributes $\mathbf{X}$, and the adjacency matrix $\mathbf{A}$.

\noindent \textbf{Zero-shot Graph OOD Detection}.
Only the names of ID classes, node attributes $\mathbf{X}$, and the adjacency matrix $\mathbf{A}$ are available, with no access to any labels for ID or OOD nodes. This setting is consistent with the zero-shot OOD detection paradigm in the image domain.

\section{Methodology}
\label{sec:methodology}

\textbf{Overview}.
We address the zero-shot graph OOD detection problem by adapting GFM—a closed-world zero-shot classification method—to the open-set OOD setting. Notably, the original GFM model lacks mechanisms specifically designed for OOD detection. To enable OOD detection, we introduce all label names to the GFM and directly use its output logits for zero-shot OOD detection (\S \ref{subsec:all labels}). However, despite its effectiveness, the true OOD label space is often inaccessible in real-world settings. Therefore, we investigate zero-shot OOD detection with only ID labels as input (\S \ref{subsec:ID labels}). 
This approach, however, suffers from limited expressiveness: it infers OOD status solely from the absence of strong affinity with ID labels. Without any explicit semantic reference for OOD, the model struggles to identify semantically ambiguous or borderline cases.
To address this, we propose prompting the LLM to generate a set of pseudo-OOD labels based on the initially unlabeled data, enabling the model to distinguish OOD samples in a more nuanced and fine-grained manner. Our method detects OOD nodes by evaluating their affinities with both ID and pseudo-OOD labels (\S \ref{sec::Pseudo-OOD}).
\vspace{-0.1in}

\begin{figure*}[!t]
   \begin{center}
   \includegraphics[width=1\linewidth]{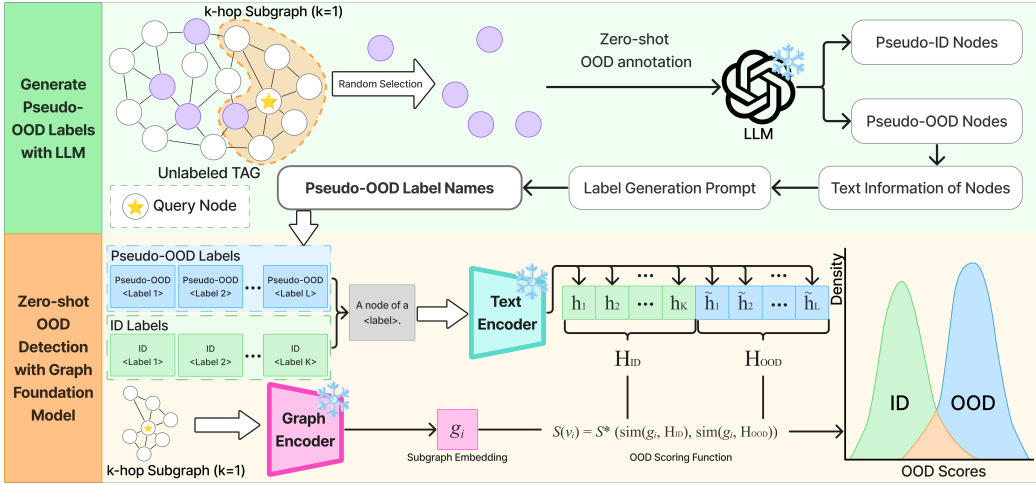} %
   \vspace{-0.2in}
    \end{center}
 \caption{\method overview. Initially, the entire graph is unlabeled, and the labels of all nodes are unknown. \method leverages an LLM to generate semantically informative pseudo-OOD labels from unlabeled data, enabling the GFM to capture nuanced semantic boundaries between ID and OOD classes without requiring any labeled nodes.
}
\vspace{-0.1in}
\label{fig:framwork}
\end{figure*}

\subsection{Zero-shot Inference with Graph Foundation Model}
\noindent \textbf{Graph Encoder}. 
For each node \( v_i \), we construct its corresponding induced subgraph \( G_i = (\mathbf{P}_i, \mathbf{X}_i, \mathbf{A}_i) \), where \( \mathbf{P}_i \) denotes the positional encodings, \( \mathbf{X}_i \) is the feature matrix of the subgraph (e.g., the \( k \)-hop neighborhood centered at node \( v_i \)), and \( \mathbf{A}_i \) is the adjacency matrix of the subgraph. The subgraph is then encoded using a Graph Transformer \( g_\theta \), such as GraphGPS~\cite{rampavsek2022recipe}, as follows:
\begin{equation}
    g_i = g_\theta(\mathbf{P}_i, \mathbf{X}_i, \mathbf{A}_i), \label{eq:graph-encoding}
\end{equation}
where \( g_i \in \mathbb{R}^{1 \times d} \) denotes the graph-level embedding obtained via mean pooling over node representations. This results in a fixed-dimensional embedding \( g_i \) for each input node \( v_i \).

\noindent \textbf{Text Encoder}. 
Assuming the graph contains \( K \) classes, we can construct a text description for each class label. For instance, in the context of a citation network, a label-associated sentence can be formulated as:  
``This paper belongs to \{class\}.''  
We collect all $K$ such label sentences and denote the set as $\mathcal{T} = \{T_1, T_2, \dots, T_{K}\}$. To obtain their embeddings, we apply MiniLM~\cite{wang2020minilm} as follows:
\begin{equation}
    h_k = h_\phi(T_k), \label{eq:text-encoding}
\end{equation}
where \( h_\phi \) denotes the MiniLM encoder and $h_k \in \mathbb{R}^{1 \times d}$ is the embedding of each label sentence.

\noindent \textbf{Closed-world Zero-shot Inference on Classification Task}.
The following inference formulation enables zero-shot classification without requiring access to labeled nodes by aligning node embeddings with semantically meaningful label embeddings generated by a text encoder:
\begin{equation}
\hat{y}_i \;=\;
\underset{k\in\{1,\dots,K\}}{\operatorname*{arg\,max}}
\operatorname{sim}\!\bigl(g_i,h_k\bigr),
\end{equation}
where $\text{sim}(g_i, h_k)$ denotes the similarity between the node embedding $g_i$ and the text embedding $h_k$ corresponding to the $k$-th label. The label sentence with the highest similarity is selected as the predicted label for node $v_i$.
However, this approach assumes a closed-world setting, where the set of possible classes is fixed and known in advance. As a result, the model is limited to classifying each node into one of the predefined ID classes. In subsequent sections, we relax this constraint to explore zero-shot open-world scenarios, where OOD nodes may exist beyond the known label set.

\subsection{Zero-shot Graph OOD Detection with All Labels}
\label{subsec:all labels}
In this section, we provide all the label names to our framework for zero-shot OOD detection. 
Let $\mathcal{T}_{\text{ID}} = \{T_1, T_2, \dots, T_{K_{\text{ID}}}\}$ denote the set of ID label texts, and let $\mathcal{T}_{\text{OOD}} = \{\bar{T}_1, \bar{T}_2, \dots, \bar{T}_{K_{\text{OOD}}}\}$ denote the set of OOD label texts. Their corresponding embeddings are denoted as $\mathcal{H}_{\text{ID}} = \{h_1, h_2, \dots, h_{K_{\text{ID}}}\}$ and $\mathcal{H}_{\text{OOD}} = \{\bar{h}_1, \bar{h}_2, \dots, \bar{h}_{K_{\text{OOD}}}\}$, where each embedding is computed using the same text encoder $h_\phi$ defined in Eqn.~\ref{eq:text-encoding}, with $h_k = h_\phi(T_k)$ and $\bar{h}_k = h_\phi(\bar{T}_k)$.
We define the OOD score for node $v_i$ using its embedding $g_i$ and both ID and OOD label embeddings as follows:
\begin{equation}
S(v_i) = S^*(\text{sim}(g_i, \mathcal{H}_{\text{ID}}), \text{sim}(g_i, \mathcal{H}_{\text{OOD}})),
\label{eq:ood-score-combined}
\end{equation}
where $\text{sim}(g_i, \mathcal{H})$ denotes the set of similarity scores between the node embedding $g_i$ and the embeddings in the label set $\mathcal{H}$, and $S^*$ is a fusion function that computes the OOD score based on these similarities. For example, we can normalize all similarity scores (from both ID and OOD labels), and define the OOD score of node $v_i$ as one minus the sum of the normalized similarity scores between $g_i$ and the ID label embeddings:
\begin{equation}
S(v_i) = 1 - \sum_{k=1}^{K_{\text{ID}}} \text{sim}(g_i, h_k),
\label{eq:ood-score-id-only}
\end{equation}
where $K_{\text{ID}}$ is the number of ID classes, and we use the cosine similarity function in this paper. A higher OOD score indicates lower alignment with any known ID label.

\subsection{Zero-shot Graph OOD Detection with ID Labels}
\label{subsec:ID labels}
Although the previous method is effective, it has a fundamental limitation: it requires the label names of OOD classes.  
Therefore, we investigate how to perform zero-shot OOD detection using only the ID label names. 
A simple way to achieve this is to compute the similarity scores between the query node's embedding $g_i$ and only the ID label embeddings, and then normalize these ID similarity scores. These scores can serve as the output logits of an ID classifier, similar to those used in supervised graph learning methods. Therefore, we can apply any post-hoc OOD detector—such as MSP \cite{hendrycks2016baseline} or the Energy score \cite{liu2020energy}—to these ID similarity scores to perform OOD detection. However, in this setting, the model has no explicit notion of what constitutes an OOD class; it can only identify OOD nodes based on their weak affinity to ID labels. 

To mitigate this limitation, a straightforward approach is to use a set of negative prompts that express semantic exclusion—such as ``This node does not belong to [ID class]'' or ``This is not about [ID label]''~\cite{wang2023clipn}—to serve as pseudo-OOD label sentences. These prompts aim to capture negation semantics and enable the model to infer OOD-ness by contrasting a node’s similarity to ID labels versus these negative prompts. Formally, the set of negative prompts serves as the pseudo-OOD label set \( \mathcal{T}_{\text{OOD}} = \{\bar{T}_1, \bar{T}_2, \dots, \bar{T}_{K_{\text{OOD}}}\} \), which, together with \( \mathcal{T}_{\text{ID}} \), is used in Eqn.~\ref{eq:ood-score-combined} to compute OOD scores. However, we observe that this approach underperforms in practice. Negative prompts tend to introduce ambiguous or weak semantic signals, making it difficult for the model to reliably distinguish OOD nodes—especially when OOD cases are semantically close to ID classes. This motivates the need for more effective strategies to enhance the GFM's OOD awareness.

\subsection{Zero-shot Graph OOD Detection with Pseudo-OOD Labels}
\label{sec::Pseudo-OOD}
\noindent \textbf{OOD label names generation with LLM}. 
To enhance the model's OOD awareness, we aim to obtain negative labels that are semantically distant from the given ID labels but close to the actual, unknown OOD labels. 
Unlike image and text OOD detection, node-level graph OOD detection typically operates in a transductive setting, where the full set of nodes is accessible during training, but only the labels for ID classes are provided. Since all existing methods follow this setting, we propose leveraging LLMs to generate pseudo-OOD label names based on the initially unlabeled graph. As illustrated in Fig.~\ref{fig:framwork}, our approach utilizes the textual information associated with nodes to prompt the LLM to infer semantically meaningful pseudo-OOD labels without requiring access to real OOD data.
We begin by randomly sampling a small set of unlabeled nodes and prompting the LLM to perform zero-shot open-world annotation. Specifically, we provide the names of the ID classes to the LLM and ask whether each unlabeled query node belongs to any of these ID classes, based on its textual information.
The LLM is prompted to determine:
\begin{equation}
\hat{y}_i \;=\;
\operatorname{LLM}\!\bigl(v_i,\mathcal{Y}_{\text{ID}}\bigr)
\;:=\;
\begin{cases}
y\in\mathcal{Y}_{\text{ID}}, & \text{if the LLM matches }v_i\text{ to an ID class},\\[4pt]
\text{OOD},                  & \text{otherwise}.
\end{cases}
\end{equation}
If the LLM predicts that a node $v_i$ does not belong to any ID class, we treat it as a pseudo-OOD node. We then prompt the LLM to summarize the node’s textual content and generate a suitable OOD category name for it: 
\begin{equation}
    y^{\text{OOD}}_i = \text{LLM-Gen}(v_i), \quad \text{if  } \hat{y}_i = \text{``OOD''}
\end{equation}

\noindent \textbf{OOD detection in augmented label space}. 
Once a set of pseudo-OOD label names is generated by the LLM, we construct an augmented label space that includes both the original ID labels and the newly synthesized OOD labels. The generated OOD labels form the set $\mathcal{Y}_{\text{OOD}}$, and the augmented label space is defined as $\mathcal{Y}_{\text{aug}} = \mathcal{Y}_{\text{ID}} \cup \mathcal{Y}_{\text{OOD}}$. From $\mathcal{Y}_{\text{aug}}$, we obtain the corresponding label sentences $\mathcal{T}_{\text{aug}}$, each of which is encoded into a text embedding using the pretrained text encoder, as shown in Eqn.~\ref{eq:text-encoding}.
During inference, for each node, we compute its similarity to all label embeddings in the augmented space using Eqn.~\ref{eq:ood-score-combined}. A node is classified as OOD if it has low affinity to all ID label embeddings but exhibits high similarity to one or more pseudo-OOD label embeddings. This enables the model to leverage both the discriminative power of ID labels and the semantic richness of LLM-generated OOD labels, facilitating accurate and fine-grained OOD detection without requiring access to real OOD label names.

\section{Experiments}
\label{sec:experiments}
\subsection{Experimental Setup}
\label{subsec:exp setup}
\noindent \textbf{Datasets}.
We utilize the following TAG datasets, which are commonly used for node classification: Cora \cite{mccallum2000automating}, Citeseer \cite{giles1998citeseer}, Ele-Computers \cite{yan2023comprehensive}, and Wiki-CS \cite{mernyei2020wiki} (see Appendix \ref{appen:dataset} for details).
We use these datasets because we perform zero-shot OOD detection based on GraphCLIP \cite{zhu2024graphclip}, and these datasets were not used during the pretraining of this GFM. Therefore, we can effectively evaluate its zero-shot OOD detection performance on unseen data.
For each dataset, we split all classes into ID and OOD sets and the ID classes for the four datasets are shown in Appendix \ref{appen:class split}. Additionally, the number of ID classes is set to a minimum of three to perform the ID classification task. 

\noindent \textbf{Baselines}.
We evaluate \method against two categories of baselines.
(1) GNN-based OOD detection with ID supervision, which trains a GNN classifier using labeled ID nodes and then applies post-hoc scoring functions to the output logits for OOD detection. This includes MSP \cite{hendrycks2016baseline}, Entropy, Energy \cite{liu2020energy}, GNNSafe \cite{wu2023energy}, and GRASP \cite{marevisiting}.
(2) Language model-based zero-shot OOD detection, which computes similarity between textual node features and ID class names without using any labeled nodes. This includes BERT \cite{devlin2019bert}, SBERT \cite{reimers2019sentence}, DeBERTa \cite{he2020deberta}, and E5 \cite{wang2022text}. Here, we use only discriminative language models for OOD detection, as generative LLMs (e.g., GPT) cannot produce soft OOD scores (e.g., those defined in Eqn.~\ref{eq:ood-score-combined}) from their output logits; instead, they yield only hard binary predictions (e.g., yes or no).

\noindent \textbf{Settings}.
For all methods that require labeled nodes to train the ID classifier, we follow the setting in \cite{song2022learning} and use $20\times K$ labeled ID nodes for training on each dataset with $K$ ID classes. Additionally, we randomly select $10\times K$ labeled ID nodes and an equal number of OOD nodes for validation. The test set consists of 500 randomly selected ID nodes and 500 OOD nodes.
For language model-based methods, we use them to encode the text associated with nodes and the information of ID labels, respectively. We then compute similarity scores between each node embedding and the embeddings of ID label sentences. Based on the normalized similarity scores, we calculate the MSP, which is used as the OOD score (results using other OOD scores are provided in Appendix~\ref{appen:more results}).
Our method has two variants: (1) using all real labels, and (2) using both ID labels and pseudo-OOD labels to perform zero-shot OOD detection.

\noindent \textbf{Implementation details}.
For all supervised methods, we employ a two-layer GCN as the ID classifier and apply various post-hoc OOD detectors to it. Each GCN is configured with a hidden dimension of 32, a learning rate of 0.01, a dropout rate of 0.5 and a weight decay of 5\text{e-}4.
For \method, we use GPT-4o-mini to identify pseudo-OOD nodes and generate pseudo-OOD label names. In our experiments, we randomly sample 80 nodes from each of the Cora and Citeseer datasets, and 120 nodes from each of the Ele-Computers and Wiki-CS datasets. The LLM is prompted to classify each node into one of the predefined ID classes or, if it determines that the node does not belong to any ID class, to generate a new pseudo-OOD label name. During zero-shot inference, we randomly select a subset of the generated labels to construct the pseudo-OOD label set, and perform OOD detection using both the pseudo-OOD and original ID labels.
Further implementation details and the prompts used in this process are provided in Appendix \ref{appen:prompts}.
All results are averaged over 5 random seeds, and all experiments are conducted on a machine equipped with an NVIDIA GeForce RTX 4080 SUPER.

\noindent \textbf{Evaluation Metrics}.
For the ID classification task, we use classification accuracy (ACC) as the evaluation metric. For the OOD detection task, we employ three commonly used metrics from the OOD detection literature \cite{wu2023energy}: the area under the ROC curve (AUROC), the precision-recall curve (AUPR), and the false positive rate when the true positive rate reaches 95\% (FPR@95). In all experiments, OOD nodes are considered positive cases. 

\subsection{Main Results}
\vspace{-0.03in}
The OOD detection results on AUROC, AUPR and FRP@95 are shown in Table~\ref{tab:results}. From the results, we conclude the following: (1) \textbf{Access to all class labels yields state-of-the-art performance.} Exposing all class labels to \method enables it to achieve the best performance among all methods, even surpassing supervised approaches that rely on a large number of ID labeled nodes to train the ID classifier. This demonstrates the strong inherent capability of the GFM to perform zero-shot OOD detection. When provided with well-defined class semantics, the GFM can distinguish ID from OOD nodes without fine-tuning, highlighting its high potential for OOD awareness.
(2) \textbf{Using only ID labels limits zero-shot OOD detection.} For language model baselines performing zero-shot OOD detection, relying solely on ID class labels is generally inadequate for distinguishing between ID and OOD nodes. This highlights the need for either additional supervision or synthetic augmentation to more effectively capture the semantic boundary between ID and OOD classes.
(3) \textbf{Pseudo-OOD label generation enhances OOD awareness.} The introduction of pseudo-OOD labels generated by our proposed approach significantly enhances the OOD awareness of the GFM. By constructing a contrastive label space that incorporates synthetic OOD semantics, our method improves the model's ability to calibrate OOD scores and more effectively distinguish ID from OOD nodes. This approach outperforms post-hoc OOD detection methods applied to the output logits of language models. 
(4) \textbf{\method is competitive with supervised baselines—without supervision.} When using ID labels and pseudo-OOD labels, \method achieves performance comparable to that of supervised methods. However, \method requires neither labeled nodes for training nor any knowledge of OOD labels. The only information it needs is the names of the ID classes, after which it can directly perform zero-shot OOD detection for any input node.
(5) \textbf{Moderate Zero-shot ID Classification Performance.} While \method is primarily designed for OOD detection, it also performs competitively in ID classification. Although it lags slightly behind fully supervised GNN-based classifiers in terms of accuracy (ACC shown in Table \ref{tab:results})—as expected due to the absence of node-level supervision—its performance remains strong enough to support reliable OOD detection.

\begin{table*}[t]
    \centering
    \caption{Performance comparison (best OOD detection results highlighted in \textbf{bold}) of different models on ID classification and OOD detection tasks. \method-L denotes the use of pseudo-labels generated by the LLM, while \method-R denotes the use of real labels.}
    \label{tab:results}
    \renewcommand{\arraystretch}{1.5}
    \setlength{\tabcolsep}{1pt}

    \resizebox{\textwidth}{!}{
    \begin{tabular}{llcccccccccccccccc}
        \toprule
        \multirow{2}{*}{\textbf{Method}} & \multirow{2}{*}{\textbf{Models}} & \multicolumn{4}{c}{\textbf{Cora}} & \multicolumn{4}{c}{\textbf{Citeseer}} & \multicolumn{4}{c}{\textbf{Ele-Computers}} & \multicolumn{4}{c}{\textbf{Wiki-CS}} \\
        \cmidrule(lr){3-6} \cmidrule(lr){7-10} \cmidrule(lr){11-14} \cmidrule(lr){15-18}
        & & ACC $\uparrow$ & AUROC $\uparrow$ & AUPR $\uparrow$ & FPR95 $\downarrow$ 
        & ACC $\uparrow$ & AUROC $\uparrow$ & AUPR $\uparrow$ & FPR95 $\downarrow$
        & ACC $\uparrow$ & AUROC $\uparrow$ & AUPR $\uparrow$ & FPR95 $\downarrow$
        & ACC $\uparrow$ & AUROC $\uparrow$ & AUPR $\uparrow$ & FPR95 $\downarrow$ \\
        \midrule
        \multirow{5}{*}{Supervised GNN} 
        & MSP & 0.9284 & 0.7831 & 0.7893 & 0.7832 & 0.8620 & 0.7316 & 0.7232 & 0.7868 & 0.8452 & 0.6518 & 0.6644 & 0.9356 & 0.8744 & 0.7884 & 0.7849 & 0.7020 \\
        & Entropy & 0.9280 & 0.7853 & 0.7883 & 0.7740 & 0.8620 & 0.7425 & 0.7289 & 0.7392 & 0.8416 & 0.6583 & 0.6660 & 0.9236 & 0.8704 & 0.7987 & 0.7931 & 0.6720 \\
        & Energy & 0.9280 & 0.8002 & 0.8025 & 0.7664 & 0.8616 & 0.7727 & 0.7605 & 0.7016 & 0.8496 & 0.6329 & 0.6518 & 0.9352 & 0.8720 & 0.8126 & 0.8095 & 0.6742 \\
        & GNNSafe & 0.9272 & 0.8718 & \textbf{0.8802} & 0.5860 & 0.8604 & 0.7836 & 0.7581 & 0.7604 & 0.8456 & 0.7438 & 0.7689 & 0.8312 & 0.8732 & 0.8925 & 0.9039 & 0.5044 \\
        & GRASP & 0.9248 & 0.8460 & 0.8379 & 0.6020 & 0.8596 & 0.7457 & 0.7178 & 0.7844 & 0.8456 & 0.6868 & 0.7247 & 0.8608 & 0.8760 & 0.8958 & 0.8979 & 0.4532 \\
        \midrule
        \multirow{4}{*}{Language Models} 
        & BERT & 0.5328 & 0.5028 & 0.4877 & 0.9296 & 0.5992 & 0.5759 & 0.5422 & 0.9092 & 0.4916 & 0.4064 & 0.4558 & 0.9908 & 0.5236 & 0.5234 & 0.5022 & 0.8808 \\
        & SBERT & 0.8436 & 0.7381 & 0.6730 & 1.0000 & 0.8044 & 0.6823 & 0.6441 & 1.0000 & 0.5728 & 0.6045 & 0.5640 & 0.7824 & 0.7312 & 0.6622 & 0.6199 & 0.8116 \\
        & DeBERTa & 0.5920 & 0.5282 & 0.5113 & 0.9184 & 0.4512 & 0.4294 & 0.4452 & 0.9272 & 0.3068 & 0.4869 & 0.5004 & 0.9536 & 0.5564 & 0.6177 & 0.6000 & 0.8928 \\
        & E5 & 0.8112 & 0.7868 & 0.7390 & 0.5620 & 0.7700 & 0.7035 & 0.6791 & 0.7964 & 0.5952 & 0.5293 & 0.5086 & 0.8588 & 0.7004 & 0.7074 & 0.6673 & 0.7596 \\
        \midrule
        \multirow{2}{*}{Ours} 
        & \method-L & 0.8836 & 0.7573 & 0.7199 & 0.7228 & 0.7708 & 0.7963 & 0.7892 & 0.9248 & 0.7932 & 0.7265 & 0.6897 & 0.6032 & 0.8244 & 0.8204 & 0.8137 & 0.6420 \\
        & \method-R & 0.8836 & \textbf{0.8810} & 0.8542 & \textbf{0.4176} & 0.7708 & \textbf{0.8745} & \textbf{0.8820} & \textbf{0.5584} & 0.7932 & \textbf{0.8742} & \textbf{0.8646} & \textbf{0.4836} & 0.8244 & \textbf{0.9282} & \textbf{0.9257} & \textbf{0.3476} \\
        \bottomrule
    \end{tabular}
    }
\vspace{-0.2in}
\end{table*}

\vspace{-0.05in}
\subsection{Visualization of Label Space}
\vspace{-0.05in}
We plot the embedding space of class label representations on the Ele-Computers and Wiki-CS datasets, where ID labels, real OOD labels, and LLM-generated pseudo-OOD labels are visualized within the same semantic space. The pseudo-OOD labels, generated using the method described in Section \ref{sec::Pseudo-OOD}, are further clustered based on their semantic similarity. We instruct the LLM to directly output the names of the clustered labels. The corresponding prompt is detailed in Appendix \ref{appen:prompts}.
From the results in Fig.~\ref{fig:two-images}, we observe that the pseudo-OOD labels generated by LLMs lie outside the ID cluster but remain relatively close to real OOD labels, indicating that they serve as effective proxies. This provides semantic evidence that LLM-generated pseudo-OOD labels occupy intermediate and informative regions in the embedding space. These labels help expand the model’s decision boundaries in a controlled manner, thereby improving zero-shot OOD detection performance.

\begin{figure}[htbp]
    \centering
    \vspace{-0.1in}
    \begin{minipage}[c]{0.57\textwidth}
        \centering
        \includegraphics[width=0.48\linewidth]{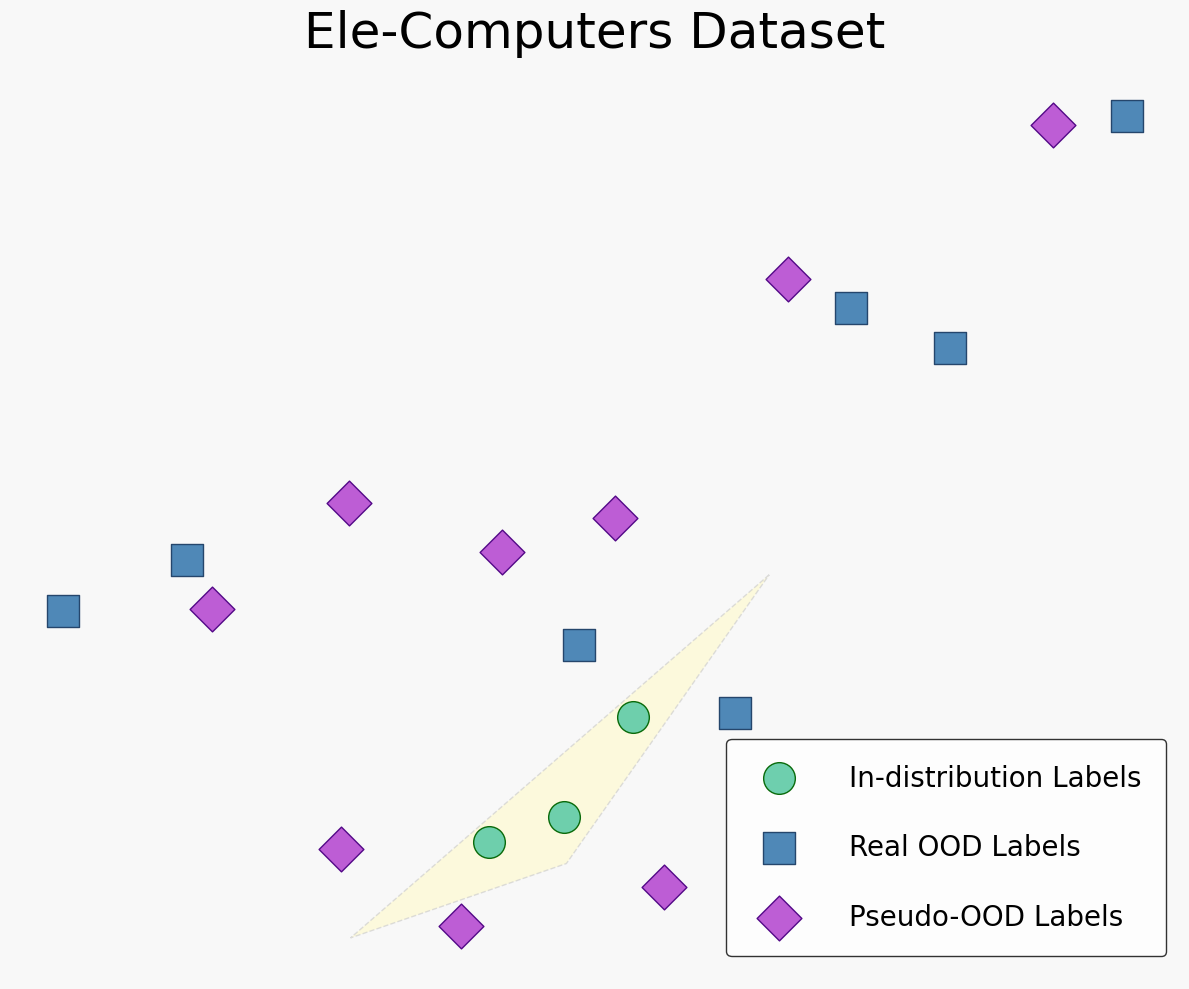}
        \hfill
        \includegraphics[width=0.48\linewidth]{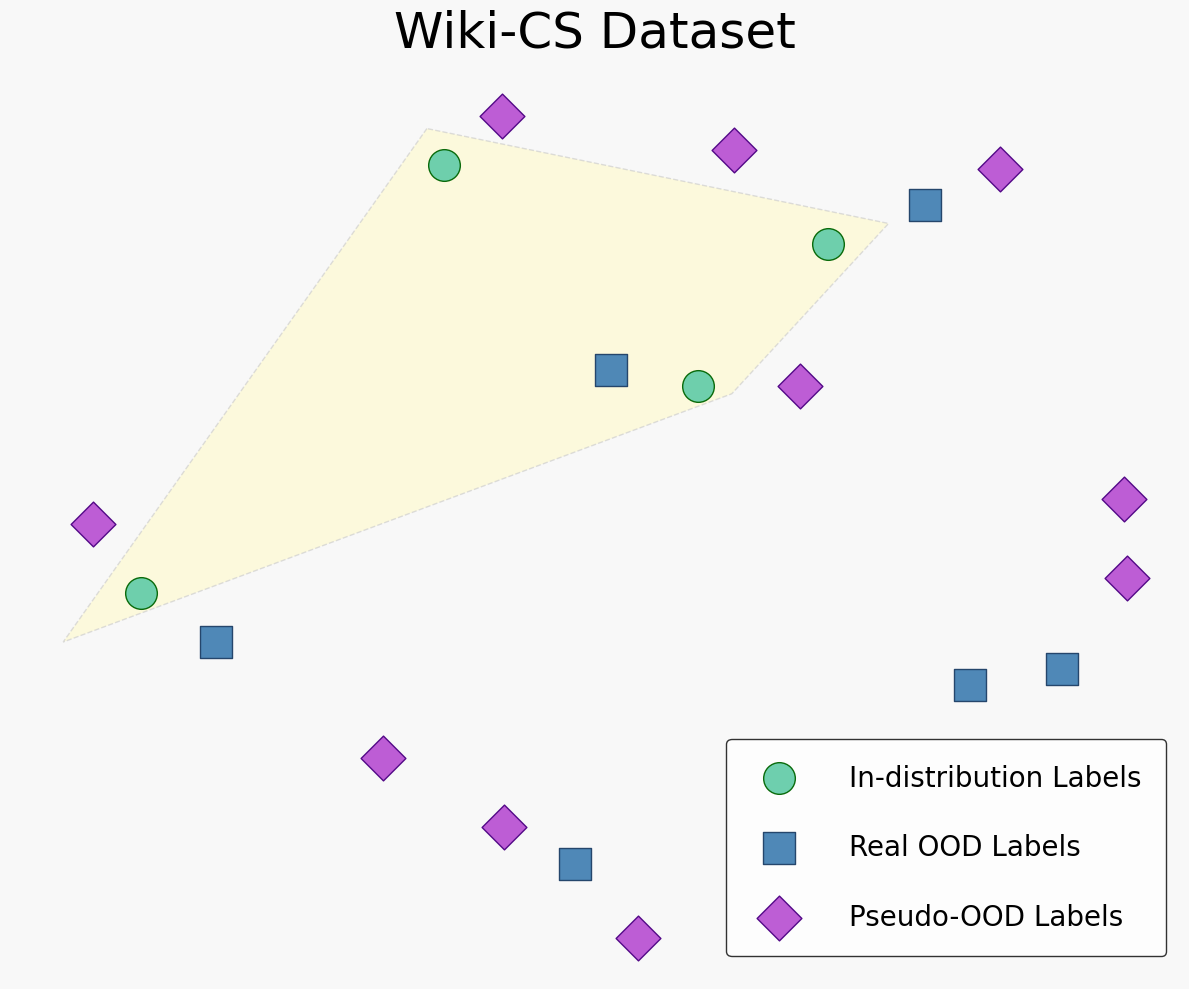}
    \end{minipage}%
    \hfill
    \begin{minipage}[c]{0.41\textwidth}
        \caption{\small
Visualization of the semantic label space.
Each point corresponds to a class label name embedded by GLIP-OOD (green for ID labels, purple for real OOD labels, and blue for pseudo-OOD labels by the LLM). The pseudo-OOD labels lie outside the ID cluster but remain semantically close to real OOD labels, highlighting the effectiveness as proxies for enhancing OOD awareness.
}
\vspace{-0.1in}
        \label{fig:two-images}
    \end{minipage}
\end{figure}

\subsection{More OOD Scores}
We use additional OOD scoring functions for zero-shot OOD detection with all labels. Following Eqn.~\ref{eq:ood-score-id-only}, the other three OOD scores are defined in Eqn.~\ref{eqn:more OOD scores}. From the results in Table \ref{tab:different OOD scores}, we observe that all four post-hoc OOD scores generally yield similar OOD detection performance, and \method remains robust across different OOD scoring functions.

\begin{align}
\label{eqn:more OOD scores}
\text{sum\_gap}(v_i) &= \sum_{k=1}^{K_{\text{OOD}}} \text{sim}(g_i, h_k^{\text{OOD}}) - \sum_{k=1}^{K_{\text{ID}}} \text{sim}(g_i, h_k) \notag \\
\text{max\_gap}(v_i) &= \max_k \text{sim}(g_i, h_k^{\text{OOD}}) - \max_k \text{sim}(g_i, h_k) \notag \\
\text{OOD\_ratio}(v_i) &= \frac{\sum_{k=1}^{K_{\text{OOD}}} \text{sim}(g_i, h_k^{\text{OOD}})}{\sum_{k=1}^{K_{\text{ID}}} \text{sim}(g_i, h_k) + \epsilon}
\end{align}

\begin{table*}[h]
    \centering
    \caption{Comparison of OOD detection performance across different OOD scoring methods.}
    \label{tab:different OOD scores}
    \renewcommand{\arraystretch}{1}
    \setlength{\tabcolsep}{1pt}
    
    \resizebox{\textwidth}{!}{
    \begin{tabular}{lccc cccc cccc cccc}
        \toprule
        \multirow{2}{*}{\textbf{Methods}} & \multicolumn{3}{c}{\textbf{Cora}} & \multicolumn{3}{c}{\textbf{Citeseer}} & \multicolumn{3}{c}{\textbf{Ele-Computers}} & \multicolumn{3}{c}{\textbf{Wiki-CS}} \\
        \cmidrule(lr){2-4} \cmidrule(lr){5-7} \cmidrule(lr){8-10} \cmidrule(lr){11-13}
        & AUROC $\uparrow$ & AUPR $\uparrow$ & FPR95 $\downarrow$ 
        & AUROC $\uparrow$ & AUPR $\uparrow$ & FPR95 $\downarrow$
        & AUROC $\uparrow$ & AUPR $\uparrow$ & FPR95 $\downarrow$
        & AUROC $\uparrow$ & AUPR $\uparrow$ & FPR95 $\downarrow$ \\
        \midrule
        sum\_gap & 0.8826 & 0.8593 & 0.4176 & 0.8756 & 0.8833 & 0.5592 & 0.8742 & 0.8645 & 0.4836 & 0.9281 & 0.9255 & 0.3476\\
        max\_gap & 0.8713 & 0.8511 & 0.4284 & 0.8706 & 0.8802 & 0.5840 & 0.8690 & 0.8595 & 0.5312 & 0.9150 & 0.9166 & 0.4284 \\
        OOD\_ratio & 0.8879 & 0.8847 & 0.4180 & 0.8776 & 0.8924 & 0.5556 & 0.8743 & 0.8677 & 0.4836 & 0.9286 & 0.9280 & 0.3476 \\
        \midrule
        sum\_ID & 0.8810 & 0.8542 & 0.4176 &
                0.8745 & 0.8820 & 0.5584
                & 0.8742 & 0.8646 & 0.4836 &
                0.9282 & 0.9257 & 0.3476 \\
        \bottomrule
    \end{tabular}
    }
    \vspace{-0.1in}
\end{table*}

\subsection{Zero-shot OOD Detection with Weak OOD Information}
In this experiment, we evaluate the performance of \method under a zero-shot setting where the model is provided with all class label names. This represents a weak OOD information scenario, where OOD label names are known but no node-level supervision is available. We compare our method against a variety of language model baselines, including BERT, SBERT, DeBERTa, and E5, which compute similarity scores between each node's text and the label names for OOD detection.

As shown in Table~\ref{tab:weak OOD}, \method consistently outperforms all language models across all datasets. This highlights the effectiveness of the GFM; even without training on node labels, it can leverage label names and graph structure to learn semantically meaningful representations, enabling robust OOD detection in a fully zero-shot setting.

\begin{table*}[h]
\vspace{-0.1in}
    \centering
    \caption{Performance comparison (best results highlighted in bold) of different models on zero-shot OOD detection when provided with all label names.}
    \label{tab:weak OOD}
    \renewcommand{\arraystretch}{1}
    \setlength{\tabcolsep}{1pt}
    
    \resizebox{\textwidth}{!}{
    \begin{tabular}{lccc cccc cccc cccc}
        \toprule
        \multirow{2}{*}{\textbf{Models}} & \multicolumn{3}{c}{\textbf{Cora}} & \multicolumn{3}{c}{\textbf{Citeseer}} & \multicolumn{3}{c}{\textbf{Ele-Computers}} & \multicolumn{3}{c}{\textbf{Wiki-CS}} \\
        \cmidrule(lr){2-4} \cmidrule(lr){5-7} \cmidrule(lr){8-10} \cmidrule(lr){11-13}
        & AUROC $\uparrow$ & AUPR $\uparrow$ & FPR95 $\downarrow$
        & AUROC $\uparrow$ & AUPR $\uparrow$ & FPR95 $\downarrow$
        & AUROC $\uparrow$ & AUPR $\uparrow$ & FPR95 $\downarrow$
        & AUROC $\uparrow$ & AUPR $\uparrow$ & FPR95 $\downarrow$ \\
        \midrule

        BERT & 0.6895 & 0.6534 & 0.7772 & 0.7378 & 0.7266 & 0.7372 & 0.5266 & 0.5023 & 0.8732 & 0.7332 & 0.7577 & 0.8268 \\
        SBERT & 0.8511 & 0.8168 & 0.4748 & 0.8662 & 0.8763 & 0.5724 & 0.7712 & 0.7490 & 0.6628 & 0.8810 & 0.8809 & 0.5056 \\
        DeBERTa & 0.6785 & 0.6528 & 0.7948 & 0.6743 & 0.6431 & 0.7884 & 0.6920 & 0.6327 & 0.7172 & 0.6694 & 0.6918 & 0.9000 \\
        E5 & 0.8636 & 0.8474 & 0.4700 & 0.8295 & 0.8302 & 0.6372 & 0.6869 & 0.6536 & 0.7312 & 0.8539 & 0.8555 & 0.5672 \\
        \midrule

        \method & \textbf{0.8810} & \textbf{0.8542} & \textbf{0.4176} & \textbf{0.8745} & \textbf{0.8820} & \textbf{0.5584} & \textbf{0.8742} & \textbf{0.8646} & \textbf{0.4836} & \textbf{0.9282} & \textbf{0.9257} & \textbf{0.3476} \\
        \bottomrule
    \end{tabular}
    }
    \vspace{-0.1in}
\end{table*}

\subsection{More OOD and ID Class Divisions}

\begin{table*}[h]
\vspace{-0.15in}
    \centering
    \caption{Comparison of OOD detection performance (best results highlighted in bold) across different models under various OOD and ID class divisions.}
    \label{tab:more division results}
    \renewcommand{\arraystretch}{1}
    \setlength{\tabcolsep}{1pt}
    
    \resizebox{\textwidth}{!}{
    \begin{tabular}{lccc cccc cccc cccc}
        \toprule
        \multirow{2}{*}{\textbf{Methods}} & \multicolumn{3}{c}{\textbf{Cora}} & \multicolumn{3}{c}{\textbf{Citeseer}} & \multicolumn{3}{c}{\textbf{Ele-Computers}} & \multicolumn{3}{c}{\textbf{Wiki-CS}} \\
        \cmidrule(lr){2-4} \cmidrule(lr){5-7} \cmidrule(lr){8-10} \cmidrule(lr){11-13}
        & AUROC $\uparrow$ & AUPR $\uparrow$ & FPR95 $\downarrow$ 
        & AUROC $\uparrow$ & AUPR $\uparrow$ & FPR95 $\downarrow$
        & AUROC $\uparrow$ & AUPR $\uparrow$ & FPR95 $\downarrow$
        & AUROC $\uparrow$ & AUPR $\uparrow$ & FPR95 $\downarrow$ \\
        \midrule
        MSP & 0.7982 & 0.7786 & 0.7436 & 0.7583 & 0.7667 & 0.8208 & 0.7305 & 0.7370 & 0.8320 & 0.7737 & 0.7739 & 0.7796 \\
        GNNSafe & 0.8819 & 0.8831  & 0.5992 & 0.8029 & 0.8142 & 0.8300 & 0.8342 & 0.8234 & 0.5432 & 0.8646 & 0.8755 & 0.6292 \\
        GRASP & 0.8891 & 0.8886 & 0.5844 & 0.7612 & 0.7540 & 0.8408 & 0.8087 & 0.7758 & 0.5656 & 0.8696 & 0.8500 & 0.5116 \\
        \midrule
        \method-R & \textbf{0.9273} & \textbf{0.9185} & \textbf{0.3484} 
                & \textbf{0.9253} & \textbf{0.9182} & \textbf{0.7016} 
                & \textbf{0.8919} & \textbf{0.8975} & \textbf{0.4952} 
                & \textbf{0.9328} & \textbf{0.9299} & \textbf{0.3008} \\
        \bottomrule
    \end{tabular}
    }
\end{table*}

To further evaluate the robustness and generalizability of \method, we experiment with an alternative set of OOD and ID class divisions across four datasets. The new ID classes are listed in Appendix \ref{appen:class split}. From the results in Table \ref{tab:more division results}, we observe that \method consistently achieves the best performance across all datasets and metrics. This demonstrates that the GFM possesses strong inherent capability for OOD detection and can adapt to different open-set configurations, thereby enhancing the reliability and robustness of zero-shot graph OOD detection.

\section{Related Works}
\subsection{Graph OOD Detection}
\vspace{-0.05in}
Node-level graph OOD detection aims at detecting nodes whose attributes or structural context deviate from the distribution of known ID classes, despite being embedded in the same transductive graph during training and inference. GNNSafe \cite{wu2023energy} demonstrates that standard GNN classifiers inherently exhibit OOD detection capabilities and introduces an energy-based discriminator trained using conventional classification loss. OODGAT \cite{song2022learning} explicitly distinguishes inliers from outliers during feature propagation and solves the problem of node classification and outlier detection in a joint framework. GRASP \cite{marevisiting} explores the advantages of propagating OOD scores and provides theoretical insights into the conditions under which this propagation is effective. Furthermore, it proposes an edge augmentation strategy with theoretical guarantees to enhance post-hoc node-level OOD detection. However, all existing graph OOD detection methods rely on a large number of labeled ID nodes for training. Our method is the first to require no labeled nodes and is the first approach for zero-shot graph OOD detection.

\subsection{Zero-shot OOD Detection}
\vspace{-0.05in}
Zero-shot OOD detection has been widely studied in the image domain \cite{wang2023clipn, esmaeilpour2022zero, ming2022delving, cao2024envisioning,liu2024cood} and text domain \cite{zhang2024your, jin2022towards}. These methods typically provide large-scale pretrained models with only the label names of ID classes and use similarity scores to distinguish ID from OOD inputs. For example, EOE \cite{cao2024envisioning} designs LLM prompts guided by visual similarity to generate potential outlier class labels tailored for zero-shot OOD detection.
ZOC \cite{esmaeilpour2022zero} extends the CLIP model to dynamically generate candidate unseen labels for each test image and introduces a novel confidence score based on the similarity between the test image and both seen and generated unseen labels in the feature space. CLIPN \cite{wang2023clipn} enhances CLIP with the ability to distinguish between ID and OOD samples by leveraging positive-semantic and negation-semantic prompts.
\cite{jin2022towards} proposes an ID-label-free OOD detection method that leverages unsupervised clustering and contrastive learning to learn effective data representations for OOD detection.
Despite these advances, zero-shot OOD detection in graph domains remains largely unexplored. This is due to the unique challenges posed by relational inductive biases and the lack of large-scale, general-purpose foundation models specifically trained for graphs. Our work bridges this gap by enabling zero-shot node-level OOD detection in graphs through the use of GFMs and language models for pseudo-OOD label generation—without requiring any labeled nodes.

\section{Conclusion}
\label{sec:conclusion}
In this work, we introduce \method, a novel framework for zero-shot graph OOD detection that leverages the capabilities of graph foundation model. Unlike existing graph OOD detection methods that rely on large numbers of labeled ID nodes, \method operates in a fully zero-shot setting, requiring only the names of ID classes and the graph structure. By prompting LLMs to generate pseudo-OOD labels, our method enables GFMs to learn meaningful semantic distinctions between ID and OOD nodes without any supervised training.
Through an extensive evaluation on four benchmark datasets, we demonstrate that when provided with all label names, \method outperforms all supervised methods. When given only ID label names, it achieves performance comparable to supervised approaches that rely on a large number of labeled ID nodes for training. Future research could explore more effective strategies for generating negative labels to further enhance the inherent OOD-awareness of increasingly powerful graph foundation models. 

\noindent \textbf{Broader Impact}.
By enhancing the OOD-awareness of graph-based systems, our method advances the development of safer and more trustworthy AI models for open-world environments. It also paves the way for future research at the intersection of foundation models, LLMs, and graph OOD detection, offering a promising step toward scalable, reliable, and label-efficient open-world graph learning. 

\noindent \textbf{Limitations}.
While our proposed method eliminates the need for both labeled nodes and a training process for graph OOD detection, it is applicable only to TAGs. Additionally, reliance on LLMs for label generation introduces concerns regarding the biases present in the pretraining of those models.

\bibliographystyle{plain}
\bibliography{references}

\clearpage
\appendix

\section*{Appendix: GLIP-OOD: Zero-Shot Graph OOD Detection with
Graph Foundation Model}

\section{Dataset}
\label{appen:dataset}

\paragraph{Cora} The Cora dataset \cite{mccallum2000automating} consists of 2,708 scientific publications, each classified into one of seven research topics: case-based reasoning, genetic algorithms, neural networks, probabilistic methods, reinforcement learning, rule learning, and theory. Nodes represent individual papers, while edges denote citation relationships, resulting in a graph structure with 5,429 edges.

\paragraph{CiteSeer} The CiteSeer dataset \cite{giles1998citeseer} comprises 3,186 scientific articles categorized into six research domains: Agents, Machine Learning, Information Retrieval, Databases, Human-Computer Interaction, and Artificial Intelligence. Each node corresponds to a paper, with features derived from its title and abstract. Citation relationships between papers define the edges of the graph.

\paragraph{WikiCS} Wiki-CS \cite{mernyei2020wiki} is designed for benchmarking graph neural networks. Each node represents a computer science article, categorized into one of ten subfields used as classification labels. Edges correspond to hyperlinks between articles, and node features are extracted from the text content of the articles.

\paragraph{Ele-Computer} Ele-Computer dataset \cite{yan2023comprehensive} consists of nodes representing electronic products, with edges capturing frequent co-purchases or co-views. Each node is labeled based on a hierarchical three-level classification of electronics. The associated text attribute is either the most upvoted user review or a randomly selected one if no such review exists. The classification task aims to assign each product to one of ten predefined categories.

\section{ID Classes}
\label{appen:class split}

\begin{table}[ht]
\caption{ID classes for different datasets.}
\label{table:split1}
\centering
\begin{tabular}{lc}
\hline
\textbf{Dataset}       & \textbf{ID class} \\ \hline
Cora                   & [0, 1, 2]      \\ 
Citeseer        & [0, 1, 2]       \\ 
Wiki-CS           & [0, 1, 2, 3]     \\ 
Ele-Computers     & [0, 1, 2]   \\ \hline
\end{tabular}
\end{table}

\begin{table}[ht]
\caption{ID classes for each dataset in another split.}
\label{table:split2}
\centering
\begin{tabular}{lc}
\hline
\textbf{Dataset}       & \textbf{ID class} \\ \hline
Cora                   & [2, 3, 4]      \\ 
Citeseer        & [2, 3, 4]       \\ 
Wiki-CS           & [2, 3, 4, 5]     \\ 
Ele-Computers     & [2, 3, 4, 5]   \\ \hline
\end{tabular}
\end{table}

\newpage
\section{More Results}
\label{appen:more results}
For language model-based methods, we encode both the node-associated text and the ID label information using the language model. We then compute similarity scores between each node embedding and the embeddings of the ID label sentences. Based on the normalized similarity scores, we derive energy and entropy values, which serve as the OOD scores. The OOD detection performance of these methods is reported in Table \ref{tab:results_language_models_cora_citeseer} and Table \ref{tab:results_language_models_computer_wikics}.

\begin{table*}[h]
    \centering
    \caption{Comparison of OOD detection performance on Cora and Citeseer across different language models using ID labels.}
    \label{tab:results_language_models_cora_citeseer}
    \setlength{\tabcolsep}{3pt} 
    {\footnotesize
    \begin{tabular}{llccccccccc}
        \toprule
        \multirow{2}{*}{\textbf{Models}} & \multirow{2}{*}{\textbf{Post-hoc}} 
        & \multicolumn{3}{c}{\textbf{Cora}} 
        & \multicolumn{3}{c}{\textbf{Citeseer}} \\
        \cmidrule(lr){3-5} \cmidrule(lr){6-8}
        & & AUROC & AUPR & FPR95 
          & AUROC & AUPR & FPR95 \\
        \midrule
        \multirow{2}{*}{BERT}    & Energy  & 0.5377 & 0.5104 & 0.9000  & 0.6340 & 0.6069 & 0.8844 \\
                                 & Entropy & 0.4963 & 0.4820 & 0.9284  & 0.5676 & 0.5314 & 0.9092 \\
        \multirow{2}{*}{SBERT}   & Energy  & 0.7605 & 0.6966 & 0.6420  & 0.7754 & 0.7317 & 0.6952 \\
                                 & Entropy & 0.7603 & 0.6891 & 0.5920  & 0.6912 & 0.6601 & 0.7632 \\
        \multirow{2}{*}{DeBERTa} & Energy  & 0.5324 & 0.5107 & 0.9324  & 0.5642 & 0.5518 & 0.9080 \\
                                 & Entropy & 0.5047 & 0.5028 & 0.9244  & 0.4465 & 0.4504 & 0.9268 \\
        \multirow{2}{*}{E5}      & Energy  & 0.7503 & 0.6895 & 0.6580  & 0.7540 & 0.7230 & 0.7636 \\
                                 & Entropy & 0.7931 & 0.7477 & 0.5584  & 0.7201 & 0.7013 & 0.7944 \\
        \bottomrule
    \end{tabular}
    }
\end{table*}

\begin{table*}[h]
    \centering
    \caption{Comparison of OOD detection performance on Ele-Computers and Wiki-CS across different language models using ID labels.}
    \label{tab:results_language_models_computer_wikics}
    \renewcommand{\arraystretch}{1.0} 
    \setlength{\tabcolsep}{3pt} 
    {\footnotesize
    \begin{tabular}{llccccccccc}
        \toprule
        \multirow{2}{*}{\textbf{Models}} & \multirow{2}{*}{\textbf{Post-hoc}} 
        & \multicolumn{3}{c}{\textbf{Ele-Computers}} 
        & \multicolumn{3}{c}{\textbf{Wiki-CS}} \\
        \cmidrule(lr){3-5} \cmidrule(lr){6-8}
        & & AUROC & AUPR & FPR95 
          & AUROC & AUPR & FPR95 \\
        \midrule
        \multirow{2}{*}{BERT}    & Energy  & 0.4356 & 0.4717 & 0.9692  & 0.4934 & 0.4964 & 0.9428 \\
                                 & Entropy & 0.4277 & 0.4682 & 0.9908  & 0.5176 & 0.4935 & 0.8780 \\
        \multirow{2}{*}{SBERT}   & Energy  & 0.5797 & 0.5431 & 0.8556  & 0.6668 & 0.5932 & 0.7228 \\
                                 & Entropy & 0.6031 & 0.5600 & 0.7816  & 0.6669 & 0.6332 & 0.7984 \\
        \multirow{2}{*}{DeBERTa} & Energy  & 0.4804 & 0.4895 & 0.9460  & 0.4777 & 0.4880 & 0.9572 \\
                                 & Entropy & 0.6538 & 0.6190 & 0.8808  & 0.6423 & 0.6218 & 0.8844 \\
        \multirow{2}{*}{E5}      & Energy  & 0.5513 & 0.5145 & 0.8692  & 0.5445 & 0.5097 & 0.8724 \\
                                 & Entropy & 0.5192 & 0.4997 & 0.8568  & 0.7240 & 0.6854 & 0.7596 \\
        \bottomrule
    \end{tabular}
    }
\end{table*}

\newpage
\section{Prompts}
\label{appen:prompts}

In our experiments, we first randomly sample 80 nodes from each of the Cora and Citeseer datasets, and 120 nodes from each of the Ele‑Computers and Wiki‑CS datasets. For each sampled node, we replace the placeholder “Object Content” in the prompt templates with its associated text, and then invoke the LLM to classify it into one of the predefined ID classes or, if necessary, generate a new class name. Optionally, we can filter out low-quality pseudo-OOD labels generated by the LLM. During zero-shot inference, we randomly select a small subset of the remaining labels to form the pseudo-OOD label set and perform OOD detection using both these pseudo-OOD labels and the original ID labels.

\begin{tcolorbox}[width=\columnwidth, colback=white, colframe=gray, title=Zero-Shot Prompt for Class Name Generation for Ele-Computers, fonttitle=\bfseries,  
]
\label{box:prompt_ele_computers}

You are a highly knowledgeable text-classification system.

You have these in-distribution (ID) classes:
\begin{itemize}
    \item \textbf{Category 1}
    \item \textbf{Category 2}
    \item \textbf{...} 
\end{itemize}

\textbf{Dataset description:} \\
\textit{Ele-Computers} dataset is extracted from the Amazon Electronics dataset.

\vspace{0.3em}
Below is the text of a \textbf{Paper}:
\begin{quote}
``\texttt{[Insert Object Content Here]}'' 
\end{quote}

Your task is to classify this Paper based on the following instructions:

\begin{enumerate}
    \item If this Paper does \textbf{not} belong to any of the listed ID classes:
    \begin{itemize}
        \item Use your general knowledge to determine the \textbf{single most appropriate class name of an electronic product} that is \textbf{not already among the ID classes listed above}.
        \item The new class should \textbf{not be too specific nor too detailed}, and it should be an electronic product \textbf{commonly seen}.
        \item You may internally summarize or reason through the text.
        \item Your final output must be \textbf{only the name} of the generated class.
    \end{itemize}
    
    \item If this Paper \textbf{does} belong to one of the listed ID classes:
    \begin{itemize}
        \item Your final output must be the \textbf{exact name} of that ID class listed above.
    \end{itemize}
\end{enumerate}

\textbf{Important Notes:}
\begin{itemize}
    \item Do \textbf{not} select any class from the ID list unless the Paper clearly fits.
    \item Your response must be a \textbf{single line} containing \textbf{only the class name} — no punctuation, no extra text.
\end{itemize}

\textbf{Answer:}
\end{tcolorbox}
\newpage
\begin{tcolorbox}[width=\columnwidth, colback=white, colframe=gray, title=Zero-Shot Prompt for Class Name Generation for Cora, fonttitle=\bfseries]
\label{box:prompt_cora}

You are a highly knowledgeable text-classification system.

You have these in-distribution (ID) classes:
\begin{itemize}
    \item \textbf{Category 1}
    \item \textbf{Category 2}
    \item \textbf{...} 
\end{itemize}

\textbf{Dataset description:} \\
\textit{Cora} dataset consists of scientific publications.

\vspace{0.3em}
Below is the text of a \textbf{Paper}:
\begin{quote}
``\texttt{[Insert Object Content Here]}'' 
\end{quote}

Your task is to classify this Paper based on the following instructions:

\begin{enumerate}
    \item If this Paper does \textbf{not} belong to any of the listed ID classes:
    \begin{itemize}
        \item Use your general knowledge to determine the \textbf{single most appropriate class name of a machine learning-related topic} that is \textbf{not already among the ID classes listed above}.
        \item The new class should \textbf{not be too specific nor too detailed}, and it should be a \textbf{commonly-seen broad topic} in the machine learning-related field — avoid rare, niche, or overly obscure concepts.
        \item You may internally summarize or reason through the text.
        \item Your final output must be \textbf{only the name} of the generated class.
    \end{itemize}
    
    \item If this Paper \textbf{does} belong to one of the listed ID classes:
    \begin{itemize}
        \item Your final output must be the \textbf{exact name} of that ID class listed above.
    \end{itemize}
\end{enumerate}

\textbf{Important Notes:}
\begin{itemize}
    \item Do \textbf{not} select any class from the ID list unless the Paper clearly fits.
    \item Your response must be a \textbf{single line} containing \textbf{only the class name} — no punctuation, no extra text.
\end{itemize}

\textbf{Answer:}
\end{tcolorbox}

\begin{tcolorbox}[width=\columnwidth, colback=white, colframe=gray, title=General Zero-Shot Prompt for Class Name Generation for Citeseer, fonttitle=\bfseries]
\label{box:prompt_general}

You are a highly knowledgeable text-classification system.

You have these in-distribution (ID) classes:
\begin{itemize}
    \item \textbf{Category 1}
    \item \textbf{Category 2}
    \item \textbf{...} 
\end{itemize}

Below is the text of a \textbf{Paper}:
\begin{quote}
``\texttt{[Insert Object Content Here]}'' 
\end{quote}

Your task is to classify this Paper based on the following instructions:

\begin{enumerate}
    \item If this Paper does \textbf{not} belong to any of the listed ID classes:
    \begin{itemize}
        \item Use your general knowledge to determine the \textbf{single most appropriate class name} that is \textbf{not already among the ID classes listed above}.
        \item You may internally summarize or reason through the text.
        \item Your final output must be \textbf{only the name} of the new class.
    \end{itemize}

    \item If this Paper \textbf{does} belong to one of the listed ID classes:
    \begin{itemize}
        \item Your final output must be the \textbf{exact name} of that ID class listed above.
    \end{itemize}
\end{enumerate}

\textbf{Important Notes:}
\begin{itemize}
    \item Do \textbf{not} select any class from the ID list unless the Paper clearly fits.
    \item Your response must be a \textbf{single line} containing \textbf{only the class name} — no punctuation, no extra text.
\end{itemize}

\textbf{Answer:}
\end{tcolorbox}

\begin{tcolorbox}[width=\columnwidth, colback=white, colframe=gray, title=Zero-Shot Prompt for Class Name Generation for Wiki-CS, fonttitle=\bfseries]
\label{box:prompt_cs_engineering}

You are a highly knowledgeable text-classification system.

You have these in-distribution (ID) classes:
\begin{itemize}
    \item \textbf{Category 1}
    \item \textbf{Category 2}
    \item \textbf{...} 
\end{itemize}

\textbf{Dataset description:} \\

WikiCS dataset is a Wikipedia-based dataset, comprising many computer science branches as classes characterized by high connectivity. Node features are extracted from the corresponding article texts.


\vspace{0.3em}
Below is the text of a \textbf{Paper}:
\begin{quote}
``\texttt{[Insert Object Content Here]}'' 
\end{quote}

Your task is to classify this Paper based on the following instructions:

\begin{enumerate}
    \item If this Paper does \textbf{not} belong to any of the listed ID classes:
    \begin{itemize}
        \item Use your general knowledge to determine the \textbf{single most appropriate class name of a topic in computer science and engineering} that is \textbf{not already among the ID classes listed above}.
        \item The new class should \textbf{not be too specific nor too detailed}, and it should be a \textbf{commonly-seen broad topic} — avoid rare, niche, or overly obscure concepts.
        \item You may internally summarize or reason through the text.
        \item Your final output must be \textbf{only the name} of the generated class.
    \end{itemize}

    \item If this Paper \textbf{does} belong to one of the listed ID classes:
    \begin{itemize}
        \item Your final output must be the \textbf{exact name} of that ID class listed above.
    \end{itemize}
\end{enumerate}

\textbf{Important Notes:}
\begin{itemize}
    \item Do \textbf{not} select any class from the ID list unless the Paper clearly fits.
    \item Your response must be a \textbf{single line} containing \textbf{only the class name} — no punctuation, no extra text.
\end{itemize}

\textbf{Answer:}
\end{tcolorbox}

\newpage
For the Ele‑Computers dataset, we cluster the generated labels into higher‑level categories using the procedure described in the following prompt. These clustered categories are then used as pseudo-OOD labels in the subsequent steps.

\begin{tcolorbox}[width=\columnwidth, colback=white, colframe=gray, title=Clustering Prompt for Category Grouping, fonttitle=\bfseries]
\label{box:prompt_clustering}

You are a highly intelligent taxonomy generation system.

You are given a list of items to be clustered:
\begin{itemize}
    \item \textbf{generated class 1}
    \item \textbf{generated class 2}
    \item \textbf{...} 
\end{itemize}

\vspace{0.5em}
\textbf{Task:} \\
cluster the above things into \{num\_cluster\} categories and specify their names, \\
requests: \\
The generated categories should be away from the below categories:
\begin{itemize}
    \item \textbf{in-distribution class 1}
    \item \textbf{in-distribution class 2}
    \item \textbf{...} 
\end{itemize}

please generate one or two sentences to describe the above \{num\_cluster\} things, respectively.

\vspace{0.5em}
\textbf{Answer:}
\end{tcolorbox}


\end{document}